\documentclass[twocolumn]{article}

\usepackage[T1]{fontenc}
\usepackage[utf8]{inputenc}
\usepackage{newtxtext,newtxmath}   
\usepackage{microtype}

\usepackage[margin=1in]{geometry}

\usepackage{graphicx}
\graphicspath{{figure/}}
\usepackage[export]{adjustbox}
\usepackage{subcaption}
\usepackage{booktabs}
\usepackage{makecell}

\usepackage{amsmath}               

\usepackage{siunitx}
\usepackage[numbers,sort&compress]{natbib}

\usepackage{hyperref}
\hypersetup{
  colorlinks = true,
  linkcolor  = black,
  citecolor  = black,
  urlcolor   = black
}

\title{YpathRAG: A Retrieval--Augmented Generation Framework and Benchmark for Pathology}

\author{%
\makebox[\textwidth][c]{%
  \parbox{0.98\textwidth}{\centering
  Deshui Yu$^{1*}$ \quad 
  Yizhi Wang$^{1*}$ \quad 
  Saihui Jin$^{2*}$ \quad 
  Taojie Zhu$^{1}$ \quad 
  Fanyi Zeng$^{1}$ \quad 
  Wen Qian$^{1}$ \quad 
  Zirui Huang$^{1}$ \quad 
  Jingli Ouyang$^{1}$ \quad 
  Jiameng Li$^{1}$ \\
  Zhen Song$^{1,3\dagger}$ \quad 
  Tian Guan$^{1\dagger}$ \quad  
  Yonghong He$^{1\dagger}$\\[3pt]
  $^{1}$Tsinghua University Shenzhen International Graduate School, China\\
  $^{2}$China Unicom Guangdong Branch, China\\
  $^{3}$Department of Network Intelligence, Peng Cheng Laboratory, China\\
  \texttt{yds24@mails.tsinghua.edu.cn, heyh@sz.tsinghua.edu.cn}\\[3pt]
  $^{*}$Equal contribution \quad $^{\dagger}$Corresponding authors
  }%
}%
}

\date{}

\begin{document}
\maketitle

\begin{abstract}
Large language models (LLMs) excel on general tasks yet still hallucinate in high-barrier domains such as pathology. Prior work often relies on domain fine-tuning, which neither expands the knowledge boundary nor enforces evidence-grounded constraints.We therefore build a pathology vector database covering 28 subfields and 1.53 million paragraphs, and present \emph{YpathRAG}, a pathology-oriented RAG framework with dual-channel hybrid retrieval—BGE\textnormal{-}M3 dense retrieval coupled with vocabulary-guided sparse retrieval—and an LLM-based supportive-evidence judgment module that closes the retrieval–judgment–generation loop. We also release two evaluation benchmarks, YpathR and YpathQA\textnormal{-}M. On YpathR, YpathRAG attains Recall@5 of 98.64\%, a gain of 23 percentage points over the baseline; on YpathQA\textnormal{-}M, a set of the 300 most challenging questions, it increases the accuracies of both general and medical LLMs by 9.0\% on average and up to 15.6\%. These results demonstrate improved retrieval quality and factual reliability, providing a scalable construction paradigm and interpretable evaluation for pathology-oriented RAG.
\end{abstract}

\section*{Introduction}

In recent years, large language models (LLMs) have achieved remarkable progress in general language understanding and generation tasks \cite{brown2020gpt3,chowdhery2022palm,touvron2023llama}. However, their application in highly specialized medical domains such as pathology still faces significant challenges. On the one hand, general-purpose LLMs often produce vague, factually incorrect, or non-professional responses due to the lack of domain-specific knowledge support \cite{ji2023survey,wang2024medicalreasoning}; on the other hand, vertically specialized diagnostic models, although performing well in single-disease recognition, tend to suffer from poor generalization and degraded answer quality in multi-turn reasoning. With the continuous evolution of medical knowledge, the maintenance and iteration costs of such models are also high. Therefore, building a pathology question-answering system that simultaneously ensures medical professionalism and scalability has become an urgent research challenge.

Retrieval-Augmented Generation (RAG) provides a promising solution to improve the factuality and reliability of LLMs in specialized domains \cite{lewis2020rag,izacard2021leveraging}. By introducing an external knowledge base, RAG enables models to retrieve relevant literature prior to generation, thereby mitigating the limitations of parametric memory. However, when applied to pathology, traditional RAG still encounters three major bottlenecks. First, the pathology knowledge system is vast yet lacks standardized corpora and evaluation benchmarks \cite{singhal2023llm4clinical,zhang2024trustworthyRAG}, making retrieval and generation performance difficult to assess consistently. Second, pathology texts are semantically complex and terminology-dense, leading single-channel retrieval to suffer from semantic drift and noise propagation. Third, factual support judgment remains difficult to automate, often resulting in hallucinations and domain mismatches in the generation process.Therefore, the pathology community urgently needs a pathology-specific, benchmarked RAG framework—built on a large structured vector corpus and coupling hybrid dense–sparse retrieval with support-aware evidence judgment—to ensure reproducible evaluation and reliable generation.

\begin{figure}[t]
  \centering
  \includegraphics[width=1\columnwidth, keepaspectratio]{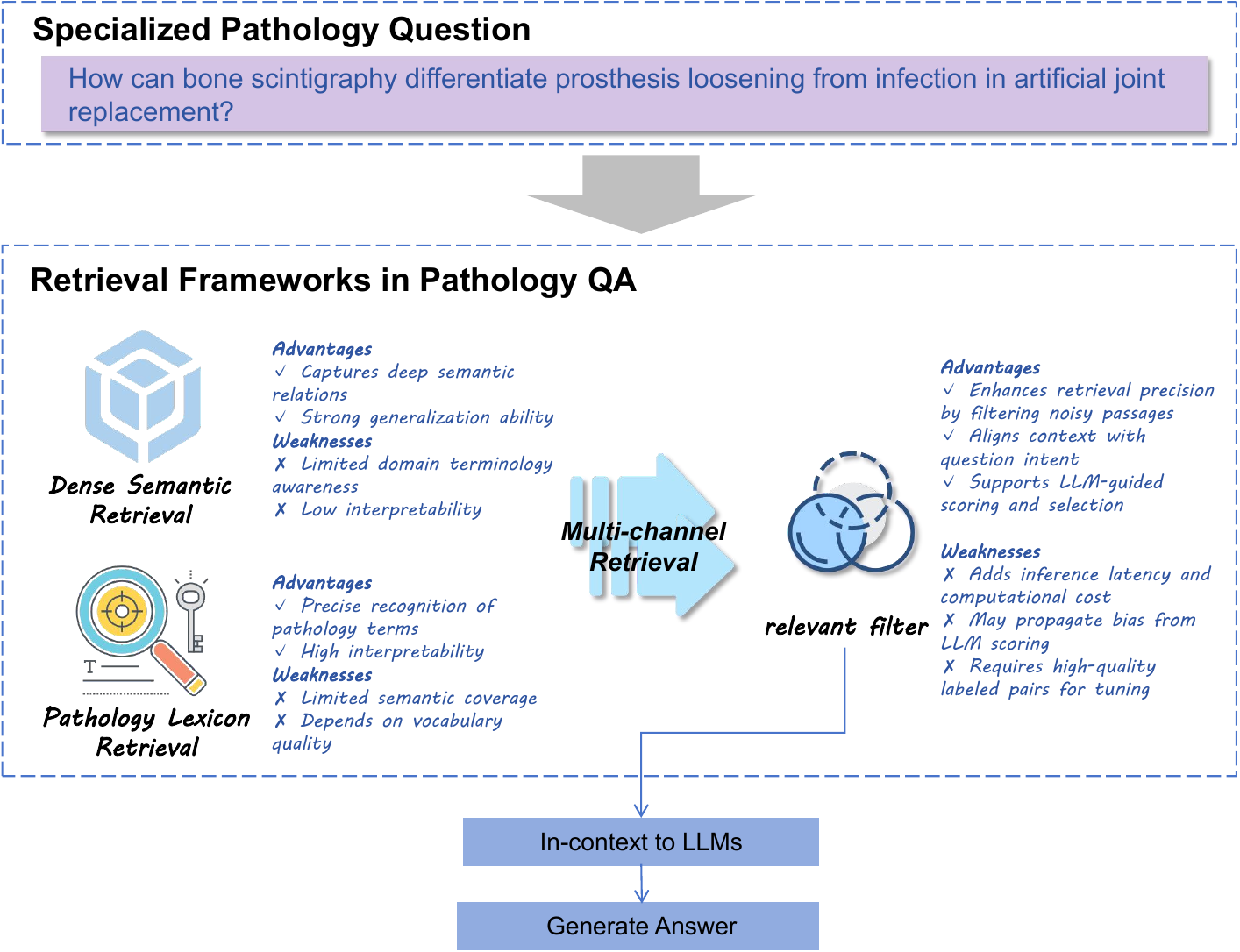}
  \caption{Overall architecture of \emph{YpathRAG}, a pathology-oriented yet diagnosis- and multimodal-capable RAG framework integrating dense retrieval, lexicon-guided sparse retrieval, and an LLM-based relevance filter for factual grounding and semantic consistency across knowledge QA, diagnostic reasoning, and multimodal QA.}

  \label{fig:motivation}
\end{figure}

\vspace{1mm}
YpathRAG is a pathology-oriented RAG framework for question answering; as shown in Fig.~\ref{fig:motivation}, it targets three persistent challenges in pathology text processing—semantic complexity, terminology density, and the difficulty of factual validation. The framework couples three synergistic modules—\emph{Dense Semantic Retrieval}, \emph{Pathology Lexicon Retrieval}, and an \emph{LLM-based Relevance Filter}—to jointly balance semantic coverage, terminology sensitivity, and factual reliability.

Specifically, the Dense Semantic Retrieval module, based on BGE-M3 embeddings, maps questions and passages into a unified vector space to capture deep semantic relationships and ensure robust cross-task generalization. The Pathology Lexicon Retrieval module leverages a domain-specific vocabulary constructed from 1.53 million pathology passages, enabling precise recognition of medical entities and long-tail terminology. Finally, the LLM-based Relevance Filter employs a large model to judge factual support and contextual alignment, filtering out semantically similar but factually inconsistent content. Together, these three modules form a multi-channel retrieval loop that enhances factual grounding, semantic coherence, and interpretability for pathology question answering.

To address the above challenges, this paper makes the following three main contributions:

\textbf{(1) Construction of a large-scale pathology vector database.}  
We systematically collected pathology textbooks, authoritative journal articles, and conference papers covering 28 subfields (e.g., pediatric pathology, reproductive pathology). Texts were extracted using OCR and denoising preprocessing. During the semantic chunking stage, we introduced an LLM-assisted chunking strategy that divides text into semantically complete segments while preserving the original wording and resolving pronoun references using contextual inference. The final database contains 1.53 million high-quality passages, providing a solid foundation for retrieval and downstream generation tasks.

\textbf{(2) Development of pathology evaluation datasets.}  
To comprehensively evaluate retrieval and question-answering performance, we constructed two benchmarks: \textbf{YpathR} and \textbf{YpathQA-M}. YpathR includes 2,440 questions automatically generated by GPT-4o from pathology literature, each paired with high-similarity positive and negative passages to test retrieval discrimination under complex semantics. YpathQA-M selects the 300 most difficult questions, forms standard answers through keyword extraction, and serves as a benchmark for assessing LLM performance on challenging pathology QA tasks. This design ensures high domain professionalism and retrieval dependency, enabling realistic evaluation of model reasoning capability in the pathology domain.

\textbf{(3) Proposal of the YpathRAG framework.}  
To address the semantic complexity and terminology density of pathology texts, we propose \textbf{YpathRAG}—a closed-loop enhancement framework following the “retrieval–judgement–generation” paradigm. YpathRAG integrates BGE-M3-based dense retrieval \cite{xu2023bge} with pathology lexicon-driven sparse retrieval to improve terminology sensitivity, and introduces an LLM-based support judgement module to filter irrelevant or weakly supported passages \cite{lewis2020rag,izacard2021leveraging,gao2022retrieval}.  

Experimental results show that YpathRAG achieves a Recall@5 of 98.64\% on YpathR, outperforming BGE-M3 \cite{xu2023bge} by 23 percentage points; on YpathQA-M, general models—GPT-4o \cite{openai2024gpt4o}, Qwen2 \cite{bai2024qwen2}, and DeepSeek-R1 \cite{deepseek2025r1}—achieve accuracy improvements of 6.2\%, 9\%, and 3.6\%, respectively, while medical-specific models (LLaVA-Med, PMC-LLaMA, HuatuoGPT) improve by 9.4\%, 15.6\%, and 9.6\%, respectively.  

These results demonstrate that YpathRAG significantly enhances retrieval precision and QA quality in pathology, aligning with recent studies on domain-specific LLM performance \cite{singhal2023llm4clinical,wang2024medicalreasoning}.

\begin{figure*}[t]
  \centering
  \includegraphics[
    max width=1\textwidth,
    max height=1\textheight,
    keepaspectratio
  ]{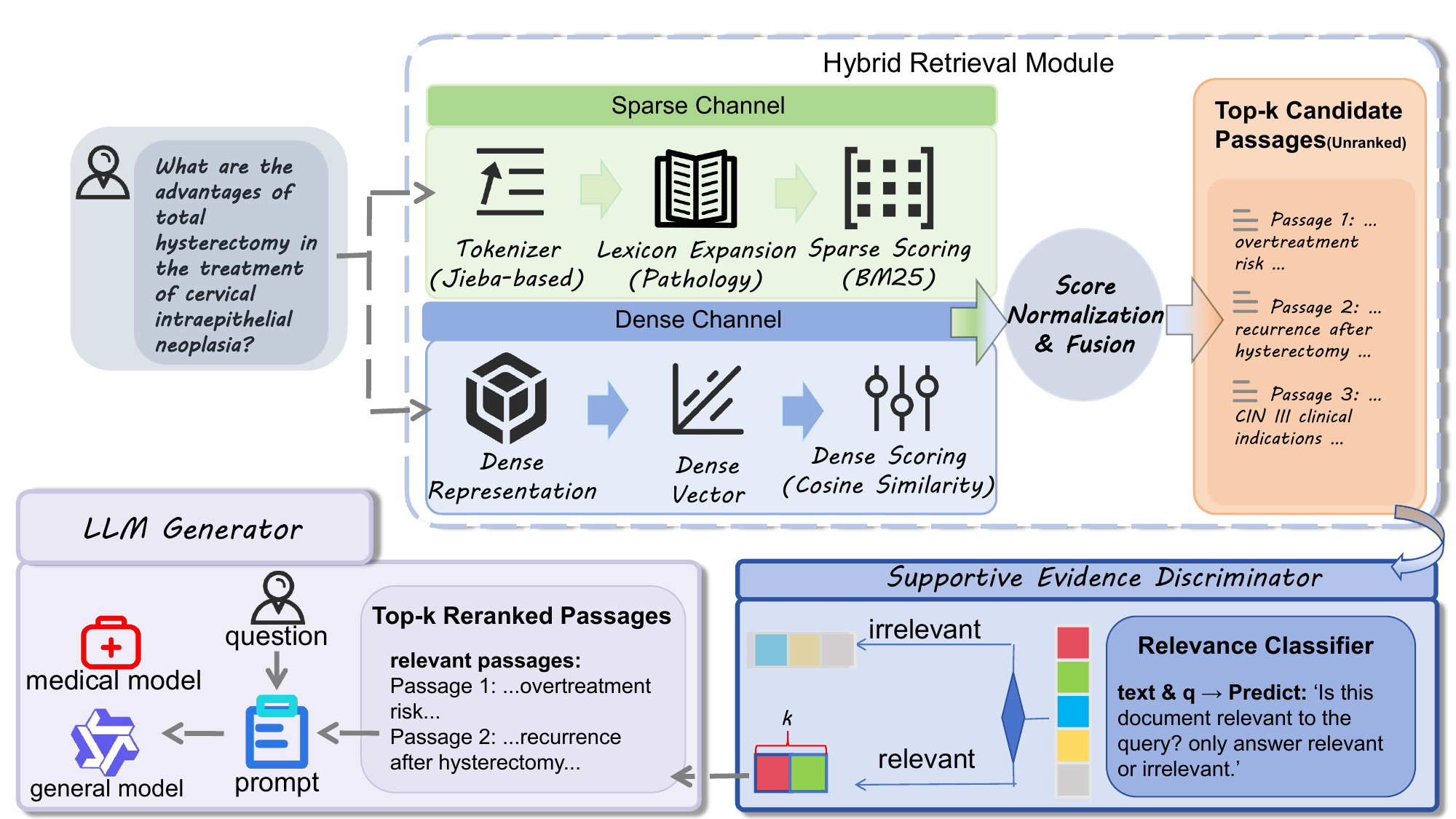}
  \caption{Overall workflow of the YpathRAG framework. The framework integrates hybrid retrieval (sparse and dense channels), supportive evidence discrimination, and a two-stage LLM generation strategy, thereby systematically enhancing retrieval and question-answering performance in pathology tasks.}
  \label{fig:framework}
\end{figure*}

The main contributions of this paper are as follows:
\begin{itemize}
    \item We present the first pathology-wide, RAG-ready vector database that demonstrably improves both general-purpose and medical LLMs on pathology benchmarks. The corpus spans 28 subfields and 1.53 million passages and is curated with LLM-assisted semantic chunking and coreference resolution to ensure consistent and complete segmentation~\cite{zeng2023pathologydataset}.

    \item We design and release two pathology-specific evaluation benchmarks—YpathR and YpathQA-M—for systematic assessment of retrieval and question-answering performance \cite{rajpurkar2018squad,zhang2024trustworthyRAG};
    \item We propose YpathRAG, a dense–sparse hybrid retrieval framework with LLM-based support judgement, which achieves consistent performance gains across multiple general and medical large language models \cite{lewis2020rag,gao2022retrieval}.
\end{itemize}

\section*{Related Work}

\textbf{Applications of large language models in medicine and pathology} \quad
In recent years, large language models (LLMs) have shown great potential in medical question answering, clinical decision-making, and reasoning tasks \cite{singhal2023llm4clinical,wang2024medicalreasoning}. Through large-scale training and domain adaptation on medical corpora, LLMs have achieved significant improvements in understanding medical terminology and professional discourse, showing promising results in certain pathology-related tasks. For instance, BioGPT \cite{luo2022biogpt} achieves leading performance in biomedical question answering, while PMC\textnormal{-}LLaMA \cite{wu2023pmcllama} substantially enhances generation quality through fine-tuning on the PubMed Central corpus.  
However, pathology, as a highly specialized subfield of medicine, continues to pose severe challenges for LLMs: first, pathology texts contain extremely dense terminology and complex semantic structures, often leading to factual errors or semantic drift \cite{ji2023survey}; second, the rapid evolution of pathology knowledge makes static corpus adaptation insufficient for capturing emerging clinical insights. As a result, LLMs tend to produce vacuous, logically shallow, or weakly interpretable answers when handling specialized pathology queries.

Retrieval–augmented generation (RAG) integrates external retrieval into the generation process to improve factuality and controllability, and has been widely studied for general NLP tasks \cite{lewis2020rag,izacard2021leveraging}. GraphRAG \cite{yang2024graphrag}, for instance, enhances inter-entity reasoning through knowledge-graph construction, improving complex reasoning performance; Qwen3\textnormal{-}Embedding \cite{qwen3embedding2025} exhibits strong generalization on MTEB, particularly in multilingual and passage-retrieval settings. However, when transferred directly to pathology, such methods face high retrieval noise due to the scarcity of domain-specific corpora and controlled vocabularies, which in turn propagates errors and degrades generation quality. This gap motivates a pathology-specific, benchmarked RAG framework with hybrid retrieval and support-aware evidence filtering.

Recent studies have begun to explore retrieval–augmented generation in medical and clinical contexts. MedRAG \cite{chen2023medrag} constructs a knowledge-enhanced system using clinical records and medical guidelines to improve the interpretability of diagnostic reasoning; BioRAG \cite{zhang2023biorag} integrates PubMed with biomedical entity linking to enhance factual consistency in biomedical QA; and Path\textnormal{-}RAG \cite{li2024pathrag} makes an early attempt to introduce retrieval augmentation for pathology report analysis in tumor diagnosis. However, these efforts largely target general clinical or image–report pipelines and still lack standardized benchmarks and frameworks for large-scale, structured retrieval over pathology text.
\begin{figure}[t]
  \centering
  \includegraphics[width=1\columnwidth]{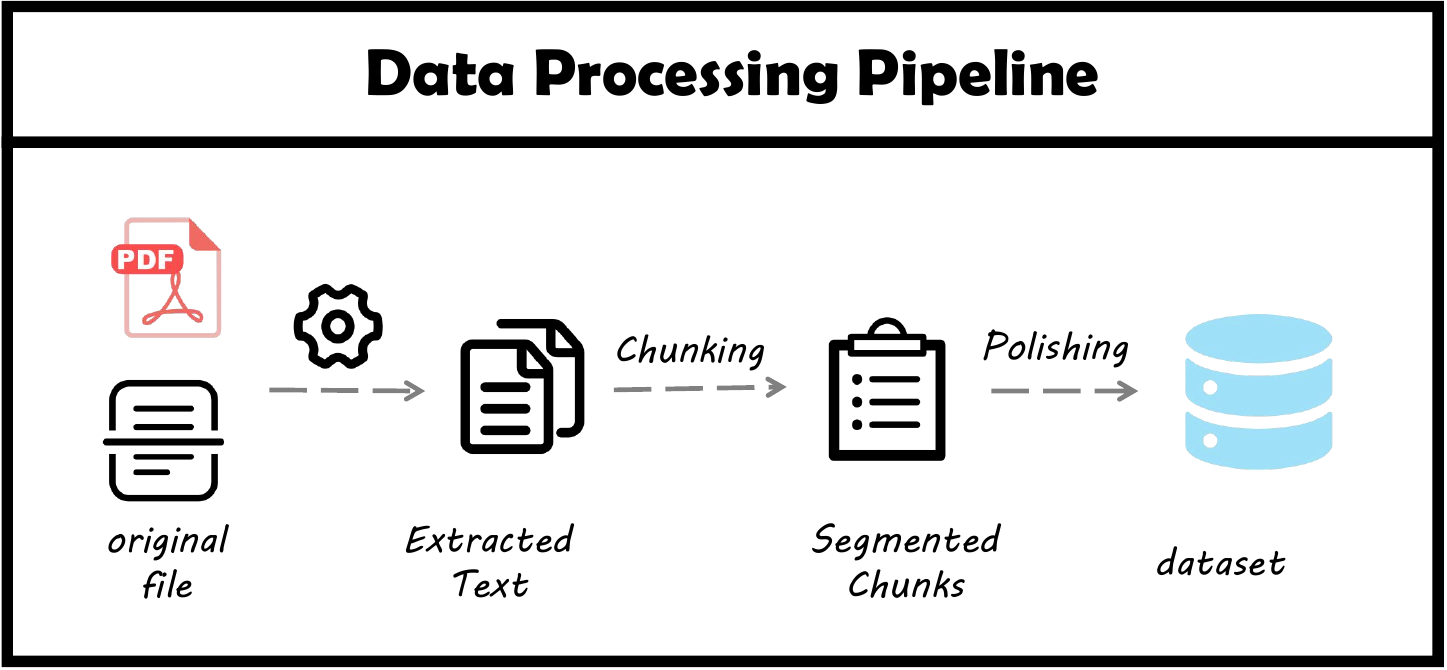}
  \caption{Data processing workflow of the pathology vector corpus. 
  The system extracts text from multi-source documents through OCR, performs semantic chunking and domain-specific refinement, and finally builds a high-quality pathology corpus containing 1.53 million semantic segments for retrieval and question answering.}
  \label{fig:datapipeline}
\end{figure}
To address these gaps, we propose YpathRAG, a retrieval–augmented framework tailored for pathology text understanding and generation. As illustrated in Fig.~\ref{fig:framework}, YpathRAG employs a \emph{hybrid retrieval module} that fuses sparse and dense channels: the sparse channel (BM25 extended with a pathology-specific lexicon) captures precise term-level matches, while the dense channel (embedding models such as BGE\textnormal{-}M3 \cite{xu2023bge}) models cross-sentence semantic associations. A \emph{score normalization and fusion} step combines the channels to form a high-confidence candidate pool, which subsequently feeds an LLM-based relevance filter for support-aware selection and reliable generation.

On top of this, YpathRAG introduces a \emph{supportive evidence discriminator} that leverages an LLM to determine whether each candidate passage provides factual support (“relevant” vs. “irrelevant”). This filtering process effectively removes semantically similar but factually inconsistent passages, substantially reducing hallucination risk. Finally, the high-confidence passages are re-ranked and passed into the LLM generator, enabling adaptive answer generation for different pathology question types (e.g., medical models vs. general models).  

This architecture achieves a closed-loop optimization of retrieval, filtering, and generation—balancing semantic comprehension, factual grounding, and contextual coherence. On the YpathR and YpathQA\textnormal{-}M benchmarks, YpathRAG achieves significant improvements in both retrieval accuracy and answer consistency over conventional RAG frameworks, establishing a reproducible structural paradigm and evaluation benchmark for pathology-domain LLM question answering.

\begin{figure*}[t]
  \centering
  \includegraphics[
    max width=1\textwidth,
    max height=1\textheight,
    keepaspectratio
  ]{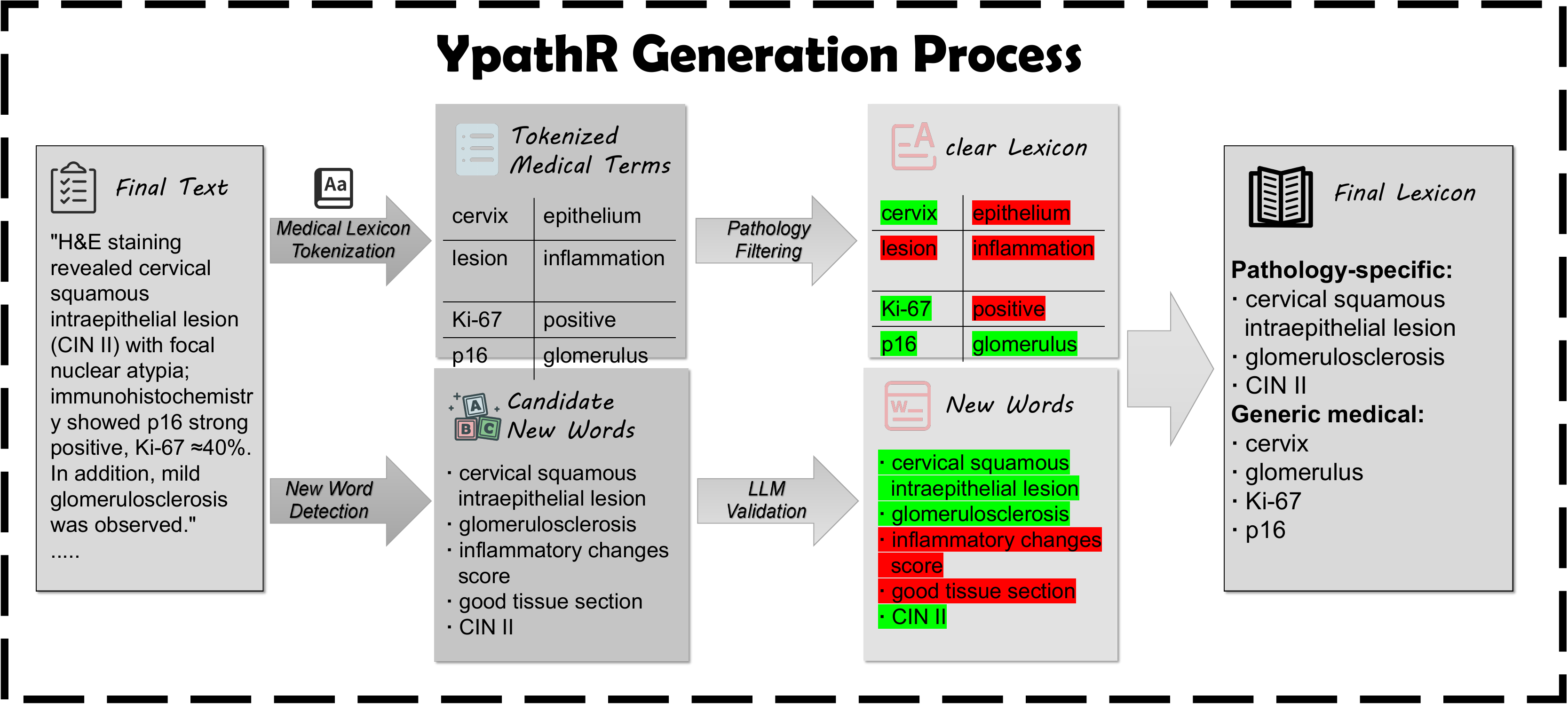}
  \caption{Construction pipeline of the YpathRAG pathology lexicon. Curated pathology texts undergo medical lexicon tokenization and new-word detection, followed by pathology filtering and LLM validation, yielding a clean lexicon that separates pathology-specific and generic medical terms. The resulting vocabulary powers the sparse retrieval channel in our hybrid RAG.}

  \label{fig:lexcion}
\end{figure*}

\section*{Pathology Corpus and Evaluation Benchmarks}

\textbf{Construction of the pathology vector corpus and preprocessing pipeline} \quad
To enable high-fidelity retrieval and reasoning in pathology, we constructed a comprehensive and structured pathology vector corpus. 
Following a multi-source literature integration strategy, we collected the top 1,000 pathology core papers from the China National Knowledge Infrastructure (CNKI), supplemented with highly cited articles, clinical guidelines, and authoritative monographs from journals such as \emph{Chinese Journal of Pathology} and \emph{Modern Pathology}, covering 28 subfields (e.g., cervical pathology, digestive system, renal pathology, endocrine pathology). 
The resulting high-quality corpus contains 1.53 million semantically coherent text segments.  

Figure~\ref{fig:datapipeline} shows the full preprocessing workflow. 
First, the system performs OCR-based text extraction from scanned and digital PDFs, handling the complex layouts and mixed fonts typical in pathology literature.  
Second, LLM-assisted semantic chunking ensures natural paragraph continuity and semantic integrity, preventing abrupt breaks or overly long spans that would degrade embedding quality.  
Third, domain agents execute specialized text normalization, including terminology standardization, symbol unification, contextual alignment, and noise removal (headers, footers, watermarks, and page numbers), followed by duplication detection and unique identifier assignment.  
Each segment is then annotated with metadata (title, year, subfield) and verified for consistency before being stored as a retrievable, vectorized knowledge unit.  

This process guarantees the structure, consistency, and traceability of the corpus, enabling YpathRAG to perform robust retrieval and factual grounding across semantically dense medical contexts.

\begin{figure*}[t]
  \centering
  \includegraphics[width=\textwidth]{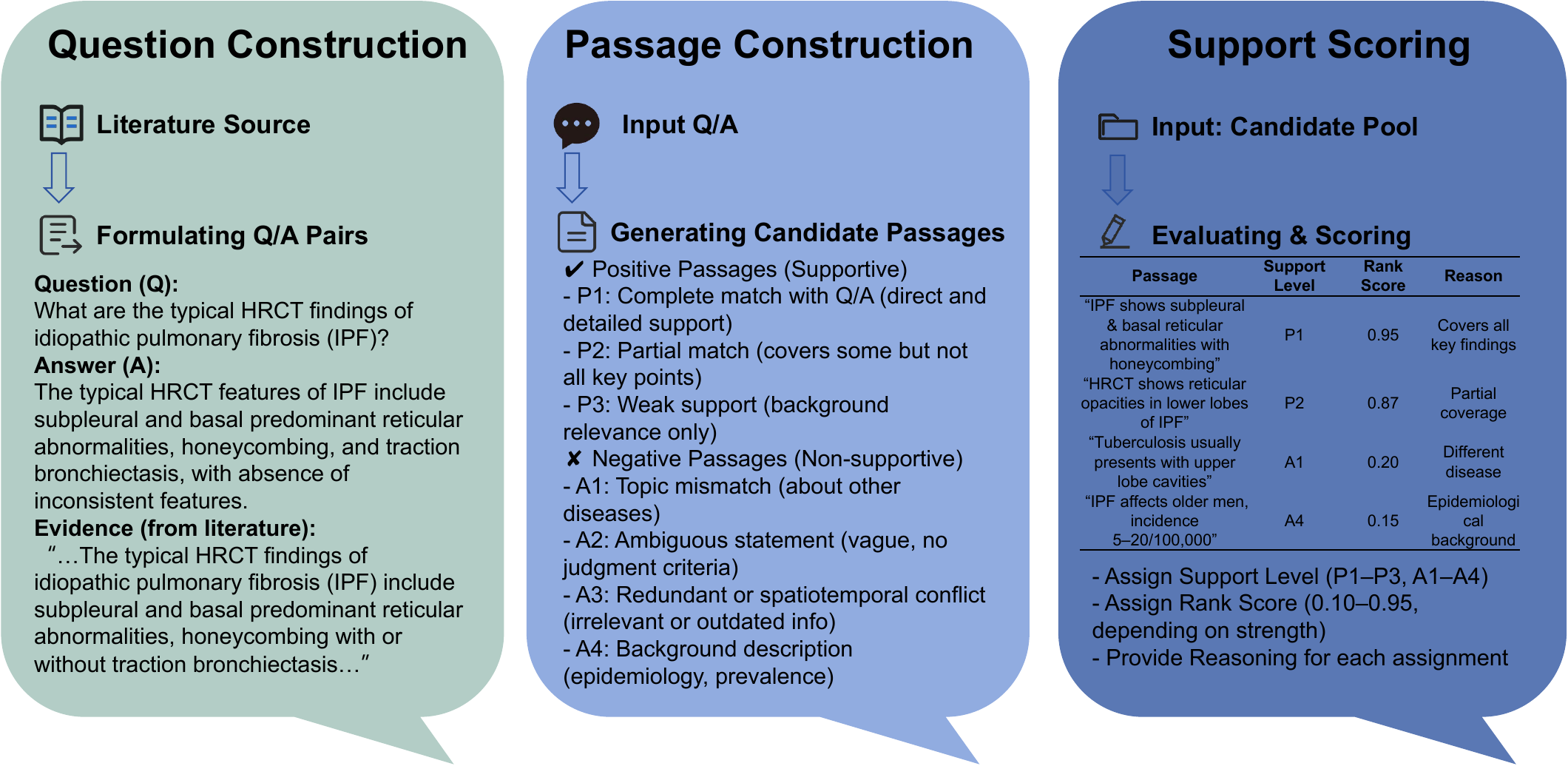}
  \caption{Construction workflow of the pathology evaluation benchmarks. 
  The process consists of three stages: (\textbf{1}) automatic generation of expert-level questions and reference answers from authoritative literature; 
  (\textbf{2}) multi-level construction of positive and negative samples (P1–P3, A1–A4) based on semantic support strength; and 
  (\textbf{3}) GPT\textnormal{-}4o-assisted support scoring and difficulty assessment to produce interpretable, high-quality benchmarks for pathology retrieval and QA.}
  \label{fig:dataset_construction}
\end{figure*}
To systematically evaluate retrieval and QA reasoning in pathology, we introduce two benchmarks: YpathR and YpathQA\textnormal{-}M. YpathR associates each query with multiple highly professional, semantically similar passages, enabling fine-grained assessment of retrieval discriminability under complex semantics. In contrast, YpathQA\textnormal{-}M is a harder subset consisting of the 300 most challenging pathology questions for evaluating large-model QA performance. Unlike open-domain QA datasets (e.g., SQuAD~\cite{rajpurkar2016squad,rajpurkar2018squad2}, Natural Questions~\cite{kwiatkowski2019natural}) and existing medical QA corpora (BioASQ~\cite{tsatsaronis2015bioasq}, MedQA~\cite{jin2021disease}, PubMedQA~\cite{jin2019pubmedqa}), our benchmarks emphasize domain specificity, high-similarity passage discrimination, and pathology-oriented question difficulty.

Currently, no large-scale benchmark systematically evaluates retrieval and QA performance in pathology. 
To fill this gap, we designed a three-stage construction pipeline (Fig.~\ref{fig:dataset_construction}) to automatically and interpretably generate questions, passages, and support annotations.

In the first stage, Question Construction: authoritative pathology literature (e.g., textbooks, reviews, and guidelines) is provided to GPT\textnormal{-}4o, which generates professional questions (\(Q\)) based on structured templates and extracts corresponding answers (\(A\)) and evidence passages. 
The questions span imaging manifestations, histopathological morphology, and underlying pathological mechanisms, ensuring comprehensive coverage of the pathology domain.  

In the second stage, Passage Construction: for each \(Q/A\) pair, a candidate pool is generated consisting of three levels of positive samples (P1–P3) and four levels of negative samples (A1–A4) according to semantic support strength.  
P1 denotes full support (direct evidence); P2 partial support (main points covered but incomplete); and P3 weak support (indirect or contextual relevance).  
A1–A4 represent topic mismatch, semantic vagueness, temporal or conditional contradiction, and non-informative context.  
This design ensures surface similarity with underlying factual divergence, approximating real retrieval discrimination challenges in pathology.  

In the third stage, Support Scoring: each candidate passage undergoes collaborative evaluation by humans and GPT\textnormal{-}4o, which annotates support levels (P1–P3, A1–A4), assigns semantic rank scores (0.10–0.95), and generates justification.  
We additionally define a question difficulty score reflecting reasoning depth and knowledge complexity.  
The final YpathR dataset includes 2,440 questions derived from 300 pathology papers, each linked to 14 passages (positive and negative).  
Among these, the most challenging 300 questions form YpathQA\textnormal{-}M, each equipped with standard answers and keyword annotations, enabling multi-dimensional QA evaluation for LLMs.  
Empirical results show that general-purpose LLMs perform poorly on YpathQA\textnormal{-}M, confirming its difficulty and suitability for evaluating domain-specific retrieval-augmented generation systems.

\begin{figure*}[t]
  \centering
  \includegraphics[
    max width=1\textwidth,
    max height=1\textheight,
    keepaspectratio
  ]{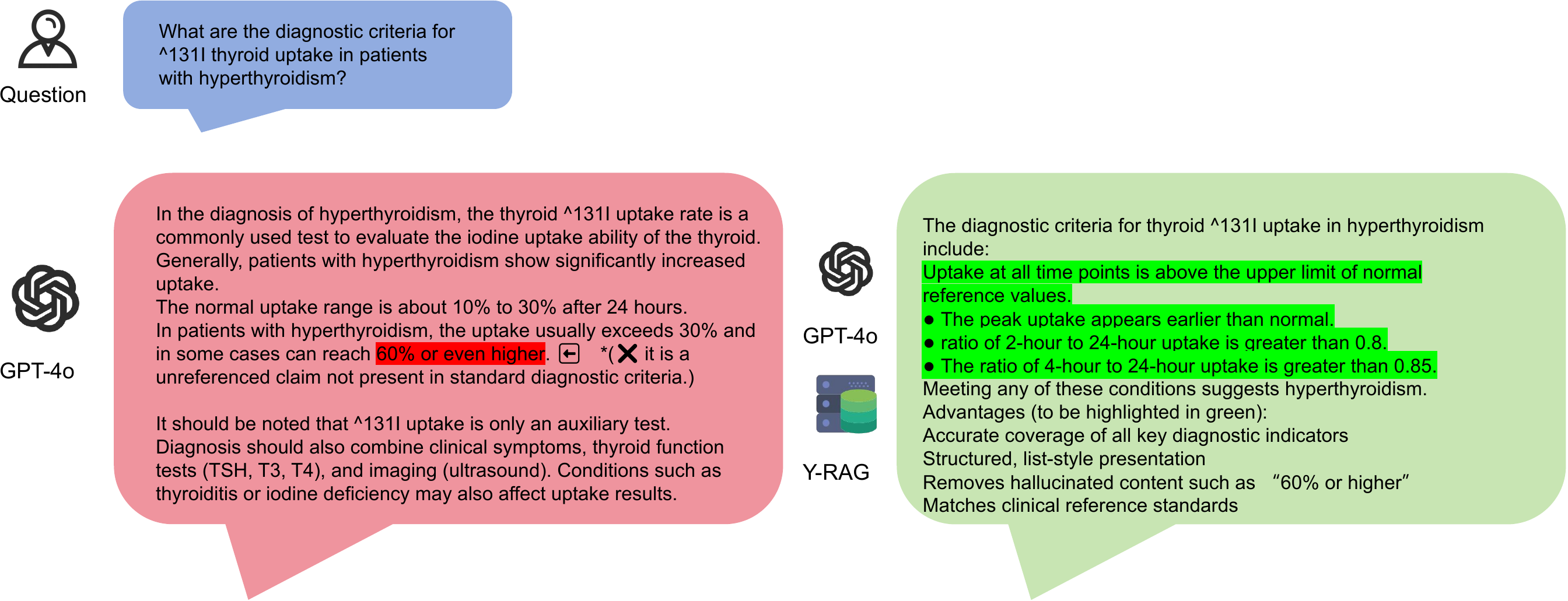}
  \caption{Case analysis of YpathRAG in pathology question answering.
  The left shows a baseline GPT-4o answer with factual hallucinations, while the right illustrates YpathRAG’s evidence-grounded generation that ensures factual consistency, structural clarity, and medical correctness.}
  \label{fig:example}
\end{figure*}

\section*{Method}

Existing RAG frameworks exhibit limited performance on pathology tasks for two primary reasons. 
First, pathology texts possess complex semantic structures and extremely high terminology density, which makes it difficult for general retrievers to capture deep semantic relations accurately. 
Second, general tokenizers and vocabularies insufficiently cover specialized medical terminology, leading to severe retrieval noise and irrelevant context \cite{ji2023survey,wang2024medicalreasoning}. 
To address these challenges, we propose the YpathRAG framework, a closed-loop process of \emph{hybrid retrieval – support filtering – enhanced generation} (Fig.~\ref{fig:framework}), designed to systematically improve retrieval precision and contextual relevance, thereby enabling both general and medical-specific large models to achieve stable and reliable performance in pathology question answering.

\textbf{Task definition} \quad
Given a question \(q\) and a document collection \(D\), the objective is to retrieve the top-\(k\) most relevant passages from \(D\) to support large language model generation. 
YpathRAG adopts a two-stage pipeline: 
(1) obtain a candidate set via hybrid retrieval; 
(2) discriminate and reorder it through support filtering; 
and finally, (3) deliver the high-confidence evidence to the generator for answer synthesis \cite{lewis2020rag,izacard2021leveraging}.

\textbf{Hybrid retrieval module} \quad
Pathology texts contain rich multi-level semantics and a dense distribution of professional terminology. 
To effectively capture both aspects, YpathRAG employs a dual-channel retrieval mechanism that integrates dense and sparse signals. 
On the dense side, BGE-M3 encodes questions and passages into dense vectors and computes their cosine similarity to model deep semantic relations \cite{xu2023bge}. 
On the sparse side, leveraging our 1.53M-segment pathology corpus, we build a domain-specific lexicon covering 28 subfields, enabling tokenization and sparse representation for precise recognition of pathology-specific entities and long-tail terms \cite{gao2021splade}. 
During fusion, scores from both channels are normalized and linearly weighted before joint ranking to form the final candidate pool. 
This design ensures a balance between semantic coverage and terminology sensitivity.

\textbf{Support filtering module} \quad
Ranking solely by semantic similarity often introduces passages that appear relevant but lack factual support, such as background descriptions, temporally misplaced statements, or thematically adjacent sentences. 
These distractors frequently trigger hallucinations during generation \cite{ji2023survey}. 
To mitigate this issue, YpathRAG introduces the Supportive Evidence Discriminator (SED), whose objective is to determine whether a passage can serve as factual evidence rather than mere semantic relevance. 
SED takes a (question, passage) pair as input and outputs a continuous \emph{support score}. 
During training, YpathR is used as supervision: P1–P3 segments are labeled as positive, and A1–A4 as negative, to train a binary relevance classifier. 
At inference, SED filters out low-support passages and fuses its scores with retrieval similarities to re-rank candidates, yielding a more factual and interpretable evidence set \cite{nogueira2020docrerank}. 
Unlike traditional re-rankers that focus on general relevance, SED emphasizes factual grounding, effectively suppressing hallucination propagation.

\textbf{Generation module} \quad
In the generation stage, YpathRAG integrates the filtered evidence with the question for controlled text generation. 
We evaluate both general-purpose models (GPT-4o, DeepSeek-R1-7B, Qwen2.5) and medical-domain models (LLaVA-Med, PMC-LLaMA, HuatuoGPT). 
Furthermore, a two-stage strategy is adopted: 
a medical-domain model first drafts a professional answer, which is then refined by a general model with enhanced coherence, factual completeness, and readability \cite{singhal2023llm4clinical,zhang2024trustworthyRAG}. 
This “domain-draft + general-refine” collaboration substantially improves both accuracy and presentation quality.

\textbf{Case study and mechanism verification} \quad
To visualize the supportive enhancement mechanism of YpathRAG, we present a representative example concerning the diagnostic criteria of $^{131}\mathrm{I}$ uptake rate (Fig.~\ref{fig:example}). 
The left side shows the baseline GPT-4o output, which, despite being fluent, introduces hallucinated thresholds (e.g., “$\geq$\SI{60}{\percent} or higher”) and mixes unrelated diagnostic indicators. 
Such factual deviations are common when general RAG systems lack a validation mechanism.  

On the right, the YpathRAG output demonstrates factual alignment and structural clarity. 
Through hybrid retrieval and support filtering, the framework extracts authoritative evidence directly describing diagnostic criteria (e.g., “uptake rates at all time points exceed normal upper limits,” “early peak uptake,” “2h/24h ratio > 0.85”) and synthesizes a structured, medically consistent answer. 
The resulting response conforms to clinical references in indicator coverage, threshold precision, and logical organization.  

This case illustrates three key advantages of YpathRAG:  
(1) suppression of hallucination propagation through pre-generation filtering;  
(2) enhanced factual granularity and structural coherence via evidence-grounded prompting;  
(3) improved balance between factuality, fluency, and domain authority through general–medical model collaboration.

\section*{Experiments}

\textbf{Experimental setup} \quad
We systematically evaluate Y\textnormal{-}RAG on two benchmarks: YpathR (for retrieval) and YpathQA\textnormal{-}M (for question answering).

\emph{YpathR}: Designed for retrieval performance evaluation, YpathR contains 2,440 questions with their corresponding positive and negative segments, totaling 14 candidate segments per question.

\emph{YpathQA\textnormal{-}M}: Composed of the 300 most challenging pathology questions with standard answers and reference keywords, YpathQA\textnormal{-}M is used to assess answer accuracy and consistency.

Answer quality is evaluated across four dimensions: keyword matching (Keyword), information coverage (Coverage), factual faithfulness (Faithfulness), and semantic similarity (Semantic Similarity), following multi–dimensional evaluation practices for large language models~\cite{lin2004rouge,zhang2019bertscore,sellam2020bleurt}.

\textbf{Retrieval performance comparison} \quad
Table~\ref{tab:retrieval} reports the retrieval results on YpathR. We compare four settings: the baseline BGE\textnormal{-}M3, the embedding model Qwen3\textnormal{-}Embedding\textnormal{-}8B, the reranker Qwen3\textnormal{-}Ranker\textnormal{-}8B, and our proposed Y\textnormal{-}RAG equipped with a Qwen2.5\textnormal{-}LoRA filter.

On Precision@5, Y\textnormal{-}RAG achieves a score of 0.9864, improving upon BGE\textnormal{-}M3 by 23.1 percentage points and significantly boosting the precision of top-ranked candidates.
On IOR-Positive (ordering consistency), Y\textnormal{-}RAG reaches 0.9169, clearly outperforming all baselines, demonstrating robust discrimination of semantically similar but unsupported passages.
Y\textnormal{-}RAG also leads in MeanRank and Hit@6, confirming stable ranking performance after dense–sparse fusion.
\begin{table}[t]
  \centering
  \caption{Performance of general LLMs on YpathQA\textnormal{-}M}
  \label{tab:general-llm}
  \setlength{\tabcolsep}{5pt}
  \resizebox{\columnwidth}{!}{%
  \begin{tabular}{l
                  S[table-format=1.4]
                  S[table-format=1.4]
                  S[table-format=1.4]
                  S[table-format=1.4]}
    \toprule
    \textbf{Model} & \textbf{Keyword} & \textbf{Coverage} & \textbf{Faithfulness} & \textbf{Semantic} \\
    \midrule
    DeepSeek\textnormal{-}R1\textnormal{-}7B                    & 0.4324 & 0.5946 & \bfseries 0.9505 & 0.6884 \\
    DeepSeek\textnormal{-}R1\textnormal{-}7B\textnormal{--}YpathRAG & 0.4681 & 0.6302 & 0.9298 & 0.6987 \\
    Qwen2:\ latest                             & 0.4758 & 0.6057 & 0.9356 & 0.7077 \\
    Qwen2:\ latest\textnormal{--}YpathRAG         & 0.5000 & 0.6398 & 0.9403 & 0.7231 \\
    GPT\textnormal{-}4o                                        & 0.4976 & 0.5795 & 0.9445 & 0.7123 \\
    GPT\textnormal{-}4o\textnormal{--}YpathRAG                   & \bfseries 0.5600 & \bfseries 0.6940 & 0.9477 & \bfseries 0.7291 \\
    \bottomrule
  \end{tabular}}
\end{table}

\begin{table}[t]
  \centering
  \caption{Performance of medical LLMs on YpathQA\textnormal{-}M}
  \label{tab:medical-llm}
  \setlength{\tabcolsep}{5pt}
  \resizebox{\columnwidth}{!}{%
  \begin{tabular}{l
                  S[table-format=1.4]
                  S[table-format=1.4]
                  S[table-format=1.4]
                  S[table-format=1.4]}
    \toprule
    \textbf{Model} & \textbf{Keyword} & \textbf{Coverage} & \textbf{Faithfulness} & \textbf{Semantic} \\
    \midrule
    LLaVA\textnormal{-}Med                       & 0.3972 & 0.4129 & 0.8225 & 0.8178 \\
    LLaVA\textnormal{-}Med\textnormal{--}YpathRAG   & 0.4870 & 0.5599 & 0.9363 & 0.8641 \\
    PMC\textnormal{-}LLaMA                       & 0.3040 & 0.4480 & 0.9000 & 0.8641 \\
    PMC\textnormal{-}LLaMA\textnormal{--}YpathRAG   & 0.3800 & 0.4587 & 0.9360 & \bfseries 0.8900 \\
    HuatuoGPT                             & \bfseries 0.4895 & 0.5600 & 0.9500 & 0.8726 \\
    HuatuoGPT\textnormal{--}YpathRAG            & \bfseries 0.4895 & \bfseries 0.6203 & \bfseries 0.9752 & 0.7262 \\
    \bottomrule
  \end{tabular}}
\end{table}

\begin{table}[t]
  \centering
  \caption{Retrieval performance comparison on the YpathR benchmark}
  \label{tab:retrieval}
  \setlength{\tabcolsep}{4pt}
  \resizebox{\columnwidth}{!}{%
  \begin{tabular}{l
                  S[table-format=1.4]
                  S[table-format=1.4]
                  S[table-format=1.4]
                  S[table-format=1.4]
                  S[table-format=1.4]}
    \toprule
    \textbf{Model} & \textbf{Precision@5} & \textbf{Hit@6} & \textbf{MeanRank} & \textbf{IOR-Global} & \textbf{IOR-Positive} \\
    \midrule
    Qwen2.5\textnormal{-}Instruct        & 0.4508 & 2.7049 & 6.7951 & 0.6294 & 0.9169 \\
    BGE\textnormal{-}M3                  & 0.7574 & 4.4385 & 4.7240 & \bfseries 0.7927 & 0.7679 \\
    Qwen3\textnormal{-}Embedding\textnormal{-}8B & 0.6926 & 3.9713 & 5.4249 & 0.7625 & 0.8654 \\
    Qwen3\textnormal{-}Ranker\textnormal{-}8B    & 0.8315 & 4.5149 & 4.5220 & 0.7221 & 0.7609 \\
    YpathRAG                & \bfseries 0.9864 & \bfseries 5.8043 & \bfseries 3.5809 & 0.7587 & \bfseries 0.9169 \\
    \bottomrule
  \end{tabular}}
\end{table}
Overall, Y\textnormal{-}RAG’s hybrid retrieval and support filtering strategy exhibits excellent domain adaptation on pathology corpora.

\textbf{Question–answering performance evaluation} \quad
Tables~\ref{tab:general-llm} and~\ref{tab:medical-llm} present results on YpathQA\textnormal{-}M for general and medical LLMs. Across all metrics, Y\textnormal{-}RAG significantly improves answer generation quality.

\emph{General models} (Table~\ref{tab:general-llm}): GPT\textnormal{-}4o, Qwen2, and DeepSeek\textnormal{-}R1\textnormal{-}7B consistently benefit from Y\textnormal{-}RAG, achieving average gains of 6.2\%–9.0\%. Notably, GPT\textnormal{-}4o\textnormal{-}Y\textnormal{-}RAG achieves 0.9477 in Faithfulness, reflecting improved factual reliability.

\emph{Medical models} (Table~\ref{tab:medical-llm}): Improvements are even more pronounced for medical-specific LLMs. HuatuoGPT\textnormal{-}Y\textnormal{-}RAG attains 0.9528 Faithfulness, while PMC\textnormal{-}LLaMA and LLaVA\textnormal{-}Med exhibit 10\%–15\% gains in Coverage and Semantic similarity. These results confirm that Y\textnormal{-}RAG’s retrieval enhancement effectively bridges domain knowledge gaps and enhances the professionalism and completeness of generated answers.

\begin{figure}[t]
\centering
\includegraphics[width=1\columnwidth]{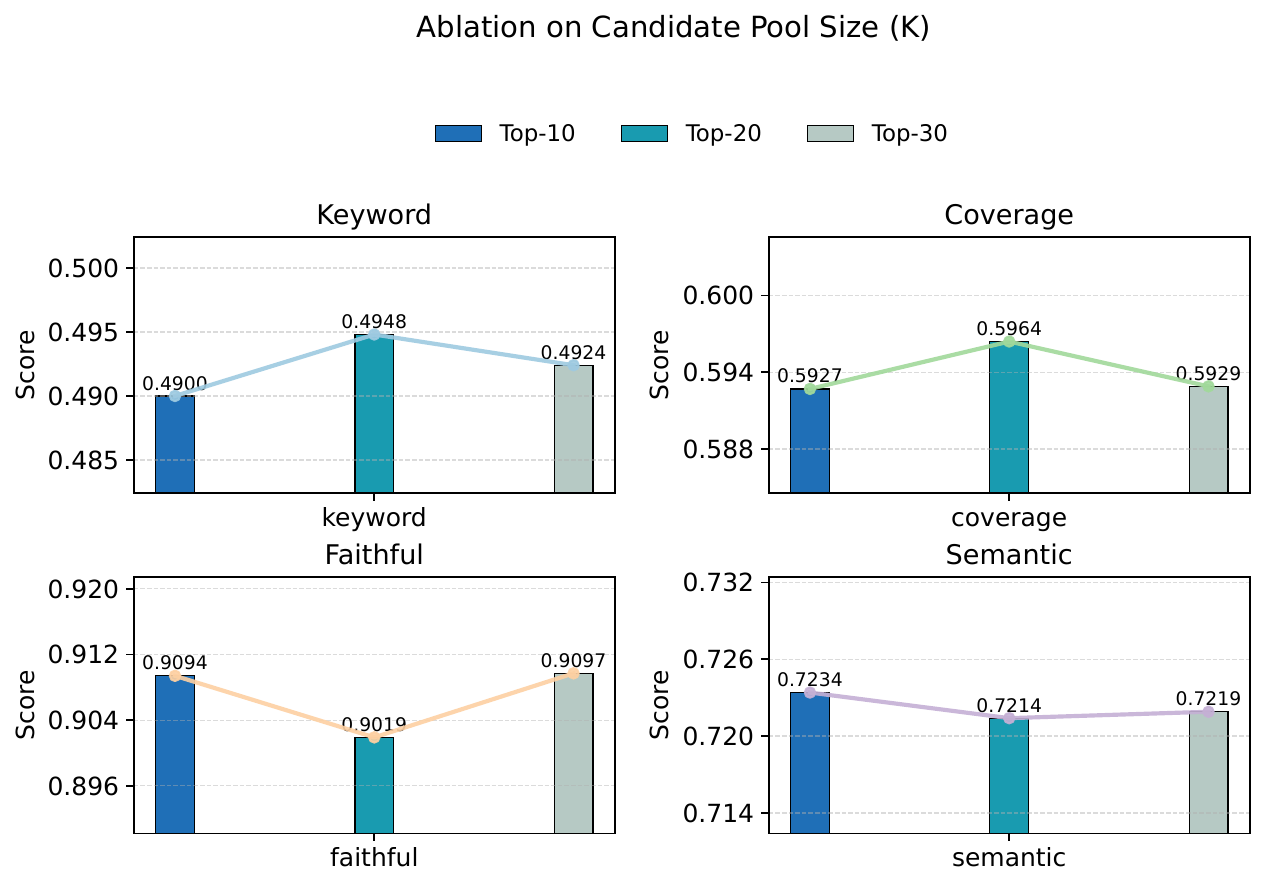}
\caption{Ablation study: effect of the number of candidate passages ($K$) on model performance.}
\label{fig:ablation_k}
\end{figure}

\begin{figure}[t]
\centering
\includegraphics[width=1\columnwidth]{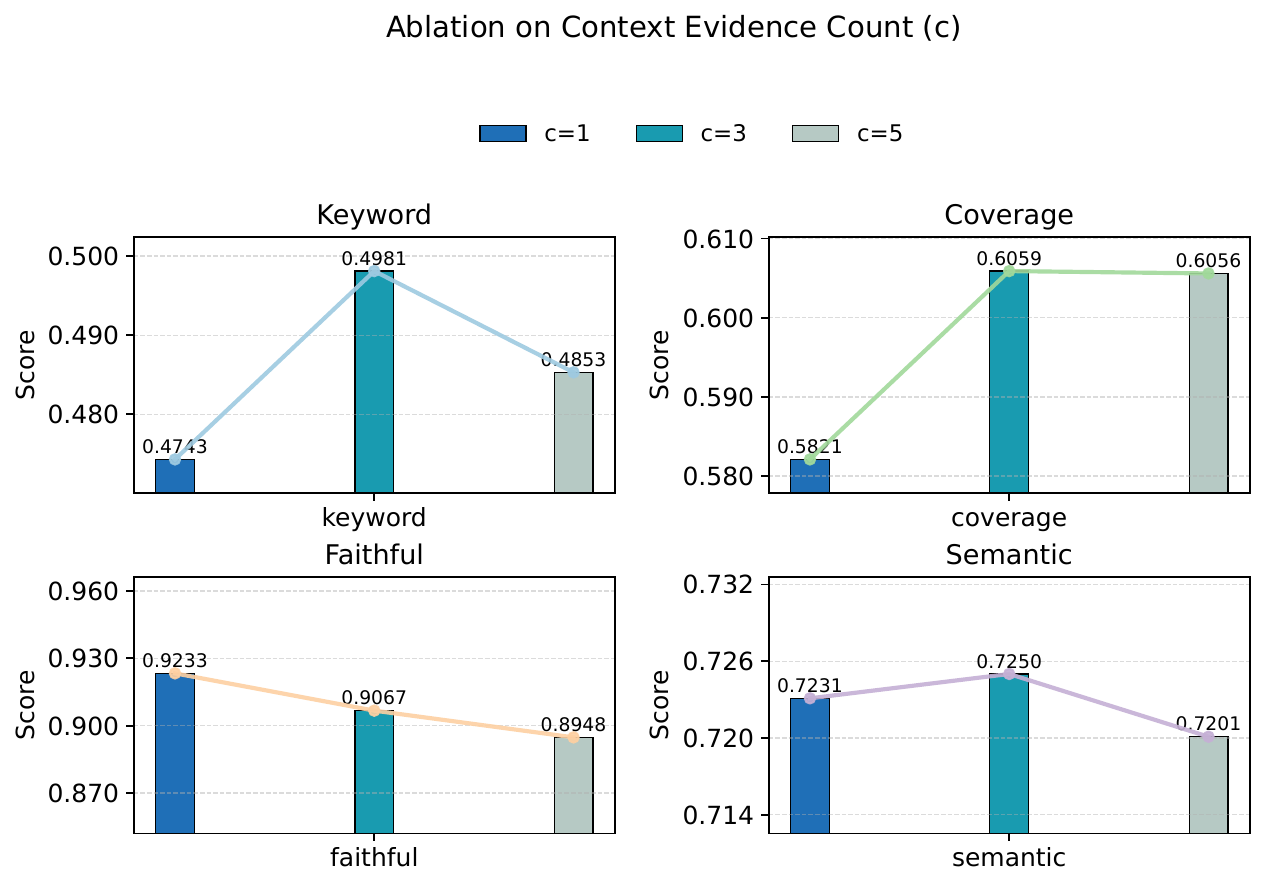}
\caption{Ablation study: effect of the number of context passages ($C$) on model performance.}
\label{fig:ablation_c}
\end{figure}

\textbf{Ablation studies} \quad
We further analyze the impact of candidate pool size and context segment number through systematic ablation experiments, as shown in Figs.~\ref{fig:ablation_k} and~\ref{fig:ablation_c}.

\emph{Candidate segment number (\(K\))}: Expanding the candidate pool from Top–10 to Top–30 increases both Coverage and Faithfulness, indicating richer evidence recall. However, when \(K > 30\), Keyword precision declines slightly due to noise from irrelevant candidates. The best balance occurs at \(K=20\).

\emph{Context segment number (\(C\))}: Performance rises notably when \(C\) increases from 1 to 3—Faithfulness and Semantic similarity improve by 2.3\% and 1.6\%, respectively. When \(C>3\), performance plateaus or slightly decreases, suggesting that excessive context can distract the generator from the main evidence. 

Overall, Y\textnormal{-}RAG achieves optimal factual completeness and semantic conciseness when the context remains within a moderate range.

\textbf{Module fine–tuning and implementation details} \quad
In the filtering module, we fine–tuned Qwen2.5\textnormal{-}Instruct\textnormal{-}7B as a binary discriminator on YpathR’s positive and negative pairs. LoRA-based parameter-efficient tuning (rank 16) was applied with a learning rate of $1\times 10^{-5}$ and batch size of 1. The fine-tuned filter achieved stable validation performance, substantially improving support discrimination and ensuring reliable evidence for answer generation.

\textbf{Summary} \quad
Y\textnormal{-}RAG significantly outperforms existing methods in both retrieval and QA, establishing new state-of-the-art results on YpathR and delivering consistent improvements across both general and medical LLMs. The pathology vocabulary and LoRA-based LLM filter are key contributors to these gains.

\section*{Conclusion}

In this paper, we introduced two benchmarks—YpathR and YpathQA\textnormal{-}M—for systematically evaluating retrieval and question-answering capabilities in pathology scenarios. Building on these, we proposed Y\textnormal{-}RAG, a pathology-specific retrieval-augmented generation framework that integrates multi-channel hybrid retrieval with a LoRA-based support filtering module. 

Experimental results demonstrate that Y\textnormal{-}RAG achieves substantial gains in retrieval precision and answer quality, outperforming strong baselines across both general and medical LLMs. This work provides a reproducible and domain-adapted paradigm for applying RAG frameworks in pathology, paving the way for reliable and interpretable medical AI systems.

\bibliographystyle{IEEEtran} 

\end{document}